%% file: main.tex
\documentclass[final]{cvpr}

\usepackage{times}
\usepackage{epsfig}
\usepackage{graphicx}
\usepackage{amsmath}
\usepackage{amssymb}
\usepackage[usenames, dvipsnames]{color}
\usepackage{xcolor}
\usepackage{pifont}
\usepackage{overpic}
\usepackage{amssymb}
\usepackage{tabu}

\usepackage[pagebackref=true,breaklinks=true,colorlinks,bookmarks=false]{hyperref}
\usepackage{float}

\input{notation}

\def\cvprPaperID{1970} 
\def\confYear{CVPR 2021}

\begin{document}


\title{Stereo Radiance Fields (SRF): \\Learning View Synthesis for Sparse Views of Novel Scenes \vspace{-5mm}}

\author{Julian Chibane$^{1,2}$\quad\quad Aayush Bansal$^2$ \quad\quad Verica Lazova$^{1,2}$ \quad\quad Gerard Pons-Moll$^{1,2}$\\
{\small ${^1}$University of Tübingen \quad ${^2}$Max Planck Institute for Informatics, Saarland Informatics Campus \quad $^3$Carnegie Mellon University}\\
{\tt \small\{jchibane, vlazova, gpons\}@mpi-inf.mpg.de, aayushb@cs.cmu.edu}}

\makeatletter
\let\@oldmaketitle\@maketitle
\renewcommand{\@maketitle}{
	\@oldmaketitle
    	\centering
	\vspace{-6mm}

}
\makeatother

\maketitle

\input{sections/Abstract}


\input{sections/Introduction}

\input{sections/Related}

\input{sections/Method}

\input{sections/Experiments}

\input{sections/Conclusions}

\noindent
\begin{minipage}{\columnwidth}
   \vspace{2mm}
   \footnotesize
	\noindent
	\textbf{Acknowledgments.}~
	\input{sections/Acknowledgements.tex}
\end{minipage}

{\small
	\bibliographystyle{ieee_fullname}
	\bibliography{references}
}
\end{document}

%% file: notation.tex
\newcommand{\mat}[1]{\mathbf{#1}}
\newcommand{\set}[1]{\mathcal{#1}}

\newcommand{\imageset}[0]{\set{I}}

\newcommand{\point}[0]{\mathbf{p}}
\newcommand{\pcolor}[0]{\mathbf{c}}
\newcommand{\pdensity}[0]{\sigma}

\newcommand{\refimg}[1]{\mathbf{I}_{#1}}
\newcommand{\feature}[1]{\mathbf{I}_{#1}(\point)}

\makeatletter
\newcommand*{\defeq}{\mathrel{\rlap{%
                     \raisebox{0.3ex}{$\m@th\cdot$}}%
                     \raisebox{-0.3ex}{$\m@th\cdot$}}%
                     =}
\makeatother

%% file: sections/Abstract.tex
\begin{abstract}
\vspace{-4mm}
Recent neural view synthesis methods have achieved impressive quality and realism, surpassing classical pipelines which rely on multi-view reconstruction. State-of-the-Art methods, such as NeRF~\cite{mildenhall2020nerf}, are designed to learn a single scene with a neural network and require dense multi-view inputs. Testing on a new scene requires re-training from scratch, which takes 2-3 days. In this work, we introduce Stereo Radiance Fields (SRF), a neural view synthesis approach that is trained end-to-end, generalizes to new scenes, and requires only sparse views at test time. The core idea is a neural architecture inspired by classical multi-view stereo methods, which estimates surface points by finding similar image regions in stereo images.
In SRF, we predict color and density for each 3D point given an encoding of its stereo correspondence in the input images.
The encoding is implicitly learned by an ensemble of pair-wise similarities -- emulating classical stereo.
Experiments show that SRF learns structure instead of over-fitting on a scene. We train on multiple scenes of the DTU dataset and generalize to new ones without re-training, requiring only $10$ sparse and spread-out views as input. We show that 10-15 minutes of fine-tuning further improve the results, achieving significantly sharper, more detailed results than scene-specific models. The code, model, and videos are available -- \url{https://virtualhumans.mpi-inf.mpg.de/srf/}. 
\end{abstract}

%% file: sections/Introduction.tex
\section{Introduction}
\label{sec:introduction}

We introduce a neural multi-view view synthesis approach which is trained end-to-end, generalizes to novel scenes, and requires only sparse views at test time (Fig.~\ref{fig:teaser}-(b)). This is in stark contrast to State-of-the-Art (SOTA) view synthesis methods like NeRF~\cite{mildenhall2020nerf}, which are trained for a specific scene and require dense multi-views to produce sharp results.

\begin{figure*}[t]
\centering
\includegraphics[width=1.0\linewidth]{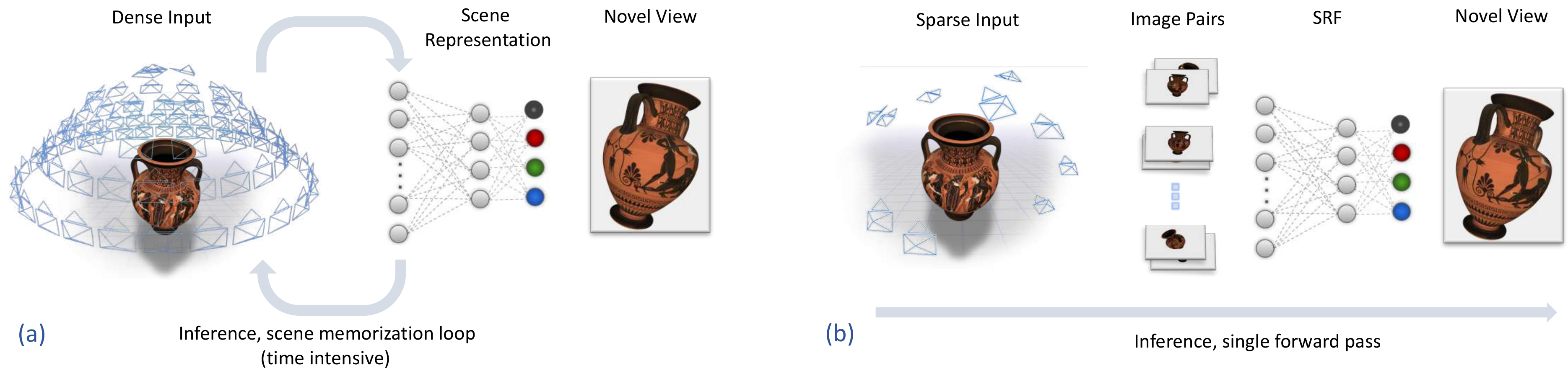}
\caption{\textbf{Pure data-driven view synthesis and SRF (ours).} 
Existing methods achieve remarkable realism representing scenes with a neural network. A model is trained specifically for a scene to synthesize high-quality novel views.
However, this requires dense views and 2 days of training per scene. In this work, we address the more challenging task of novel view synthesis from sparse and spread-out views, using a single forward pass through the network, to instantly obtain the result. 
}
\label{fig:concerns}
\end{figure*}

On one end of the view synthesis spectrum of methods, we have pure data-driven methods such as NeRF~\cite{mildenhall2020nerf}, which have shown impressive results. NeRF takes a radical data-driven approach by learning a mapping from a location and direction to the emitted radiance. This mapping is specifically trained for a scene (Fig.~\ref{fig:concerns}-(a)).  Generalization to a new scene requires retraining for 2 days and results are blurry when trained on sparse and spread-out views (Fig.~\ref{fig:teaser}-(a)). On the other end of the spectrum, popular classical image-based rendering techniques~\cite{shum2000review} use geometry~\cite{chen1993view,mcmillan1997image,seitz1996view,shade1998layered}. These approaches warp pixels to the desired target view via correspondences~\cite{rocco2018neighbourhood,scharstein2002taxonomy,szeliski2006image} or multi-view 3D reconstruction~\cite{schoenberger2016sfm,schoenberger2016mvs}. Consequently, these methods rely on high-quality 3D reconstruction or dense per-pixel correspondence, which requires dense multi-views. Recent work~\cite{BansalCVPR2020,riegler2020free} combines classical methods with data-driven approaches by learning to correct the warped views of classical methods. The sequential pipeline in these methods~\cite{BansalCVPR2020,riegler2020free} do not allow end-to-end learning.

We take inspiration from both classical and pure data-driven methods. Like NeRF, we also learn a neural network to predict radiance (specifically color and density). However, instead of memorizing the scene radiance at 3D locations, we use an image-based feature encoding, which allows the network to reason about scene geometry (Fig.~\ref{fig:concerns}-(b)). In classical stereo reconstruction~\cite{scharstein2002taxonomy,szeliski2006image}, correspondences across views are found by computing a similarity score. We devise an architecture, called \textit{Stereo Radiance Fields} (\textit{SRF}), which mimics the classical approach without computing explicit correspondences, but can be trained end-to-end.
A 3D point is projected to each available view to extract point-wise view features. Then view features are \emph{processed in pairs} by a bank of filters, which emulate correspondence finding in classical methods (Fig.~\ref{fig:method_intuition}).  
The resulting matrix of pair-wise scores is further processed with a Convolutional Neural Network~\cite{lecun2015deep} (CNN), which agglomerates information from the available views to predict the desired radiance at that point. 

Our experiments demonstrate that incorporating multi-view reconstruction ideas within the architecture significantly boosts generalization ability. 
When training on a \emph{single scene} and testing on a \emph{new scene}, SRF can produce reasonable results. This indicates that the network does not memorize the scene, but learns to reason about structure. When trained on multiple scenes (100 or more), SRF can generalize to novel scenes, even when only $10$ sparse and spread-out views are available as input. Further improvements can be obtained by fine-tuning on the $10$ views (Fig.~\ref{fig:teaser}-(c)), requiring minutes - much less than the $2-3$ days required by methods that re-train from scratch~\cite{mildenhall2020nerf,sitzmann2019scene}. SRF results are sharper, validating that multi-view reconstruction structure not only helps to generalize but also constrains the learning problem. We encourage the reader to view our results as videos available on our project page.
To summarize, our contributions are:
\\

\begin{itemize}
	\item We introduce Stereo Radiance Fields (SRF), an end-to-end, self-supervised architecture for multi-view view synthesis. We bring together insights from classical multi-view reconstruction pipelines and neural rendering approaches. 
	\item Experiments demonstrate that SRF generalize to \emph{novel scenes} given \emph{sparse and spread-out views} as input. Further, fine-tuning a pre-trained SRF for a few minutes on test distribution improve results.
	\item We show how to combine recent paradigms into one model, often treated in isolation in novel view synthesis: SRF builds on classical multi-view \textit{3D reconstruction} and \textit{learning} from multiple scenes. 
	\item In the sparse and spread-out view setting, SRF produces much sharper results than SOTA baselines like NeRF~\cite{mildenhall2020nerf}.
\end{itemize}

%% file: sections/Related.tex
\section{Multi-View View Synthesis}
\label{sec:related}

Given $N$ camera views, our goal is to synthesize a view for a new virtual camera. This is a long-standing problem~\cite{hartley2003multiple,szeliski2010computer}. Historically~\cite{shum2000review}, the problem has been studied under three possible directions depending on the geometric information used: (1) rendering without geometry~\cite{adelson1991plenoptic,gortler1996lumigraph,levoy1996light,mcmillan1995plenoptic,ng2005light} by modelling a plenoptic function to compute intensity of light rays for a given camera at every possible angle; (2) rendering using correspondences~\cite{chen1993view,seitz1996view} which requires knowledge of positional correspondences across multi-views; and (3) rendering with explicit geometry~\cite{mcmillan1997image,shade1998layered} which requires explicit 3D information in the form of depth or point clouds. In this work, we bring together insights from neural rendering with classical reconstruction pipelines. We encourage our network to reason about correspondences across pairs of views by computing an ensemble of pair-wise scores within the network. Although we never explicitly compute correspondences, this geometric reasoning, allow us to generalize to new scenes.

\noindent\textbf{Correspondences across multi-views: } Classic approaches~\cite{chen1993view,collins1996space,hartley2003multiple,seitz1996view,szeliski1999stereo} in multi-view stereo rely on correspondences across views. In this work, we bring together the insights from classical multi-view stereo~\cite{hartley2003multiple,shum2000review} and contemporary learning-based approaches~\cite{chibane2020implicit,mildenhall2020nerf,saito2019pifu}. We use an encoder network that inputs $10$ multi-views and extracts multi-scale features~\cite{bansal2017pixelnet,saito2019pifu}. We replace classical block or feature matching with a multi-layer perceptron (MLP) which outputs an ensemble of similarity scores. Like us, recent work can do view synthesis from sparse views~\cite{BansalCVPR2020} incorporating explicit correspondences. However, explicitly computing correspondences is hard due to differences in illumination, zoom, scale, and occlusion. A scene-specific model is trained to correct artifacts. In our method, the network reasons about correspondences driven by the view synthesis loss, but they are never explicitly computed. Importantly, our model is not specific to a scene.

\noindent\textbf{Neural Rendering and Plenoptic Modeling: } State-of-the-art neural rendering~\cite{tewari2020state} approaches have enabled creation of photo-realistic visual content using deep neural networks~\cite{lecun2015deep}. There are three popular directions for multi-view view synthesis: (1) using plane-sweep stereo~\cite{collins1996space,Flynn_2016_CVPR} or  multi-plane image (MPI) representation~\cite{zhou2018stereo}. MPI-based approaches~\cite{broxton2020immersive,Flynn_2019_CVPR,Flynn_2016_CVPR,Kalantari:2016,mildenhall2019local,srinivasan2019pushing} have shown remarkable results on continuous view synthesis for small baseline shifts, but fail for large ones as it assumes accurate multi-plane imaging; (2) explicitly incorporating 3D reconstruction using SfM~\cite{schoenberger2016sfm,schoenberger2016mvs} or multi-view stereo~\cite{huang2018deepmvs} for view synthesis~\cite{aliev2019neural,choi2019extreme,Hedman:2018,Meshry_2019_CVPR,riegler2020free}. These approaches assume a reasonably dense 3D point cloud used in conjunction with a neural network for view synthesis. The role of the neural network is to correct the imperfections in the 3D reconstruction. However, these approaches struggle when the views are sparse with small overlap because explicit 3D reconstruction fails; and (3) recent approaches~\cite{li2020neural,Lombardi:2019,martin2020nerf,mildenhall2020nerf,pumarola2020d,sitzmann2019scene,xian2020space,yariv2020multiview} learn a 3D representation that can be combined with differentiable-ray marching operations to synthesize a new view. These approaches by design require scene-specific modeling. 
This restricts: (1) an instant and online visualization of a new capture because it requires $2-3$ days to train a model; and (2) utilizing large amounts of diverse visual data, which has been the driving force for progress in other areas of vision such as recognition, semantic segmentation, and detection.

Our work is deeply inspired by recent neural rendering approaches. Like NeRF~\cite{mildenhall2020nerf}, we predict radiance at continuous locations and use volume rendering to generate the target image. Instead of predicting based on point coordinates and radiance, we predict based on point image features and an ensemble of similarity functions that emulate classical stereo matching. Hence, our work brings together contemporary neural rendering with classical computer vision within an end-to-end architecture. 
SRF is similar in spirit to previous work on 3D implicit shape reconstruction, Implicit Feature Networks (IF-Nets)~\cite{bhatnagar2020ipnet,chibane2020implicit} and Neural Distance Fields (NDF)~\cite{chibane2020ndf}, where we decode occupancy or unsigned distances based on volumetric deep features computed from the input, instead of the originally proposed point coordinates~\cite{i_IMGAN19,i_OccNet19,i_DeepSDF}. Our work also shares insights with contemporary approaches~\cite{trevithick2020grf,wang2021ibrnet,yu2020pixelnerf}. Finally, our work is inspired by lifelong learning~\cite{thrun1996learning,thrun1998lifelong} that aims to learn a generic representation that can be easily adapted to a new task with a few examples. 
We learn a generic view synthesis network that readily generalizes to new scenes. Our results further improve when we adapt it to the new scene with simple fine-tuning on test examples.

\begin{figure}[t]
\centering
\includegraphics[width=1.0\linewidth]{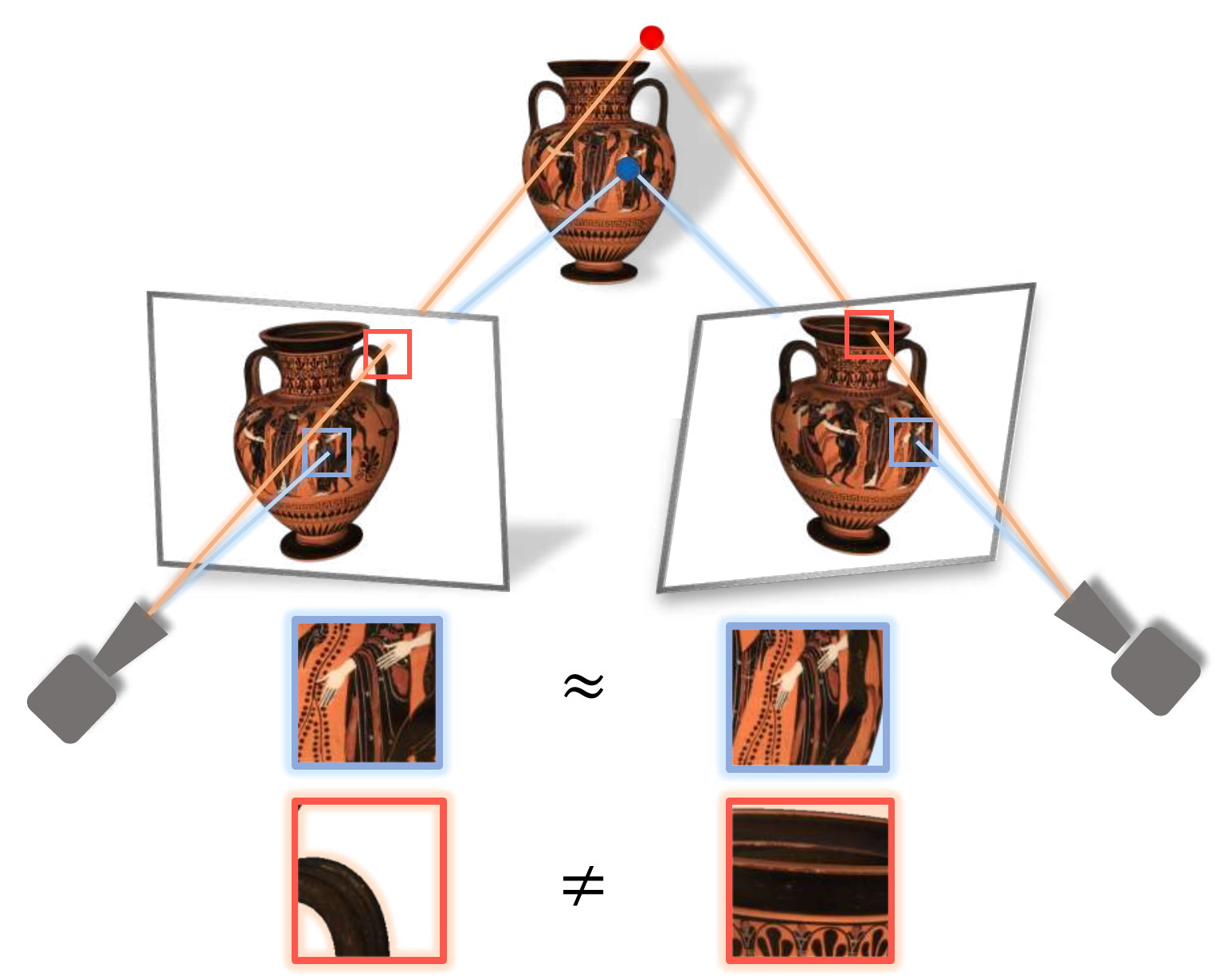}
\caption{\textbf{Intuition of our Method:} We structure our model inspired by a geometric observation: 3D points in a scene that are on a surface will project to similar-looking regions when viewed from different perspectives (blue). We call this a photo-consistent point. A point in free space, however, will not be photo consistent (red). This holds for opaque, non-occluded surface points.
}
\label{fig:method_intuition}
\end{figure}

%% file: sections/Method.tex
\begin{figure*}[t]
\centering
\includegraphics[width=1.0\linewidth]{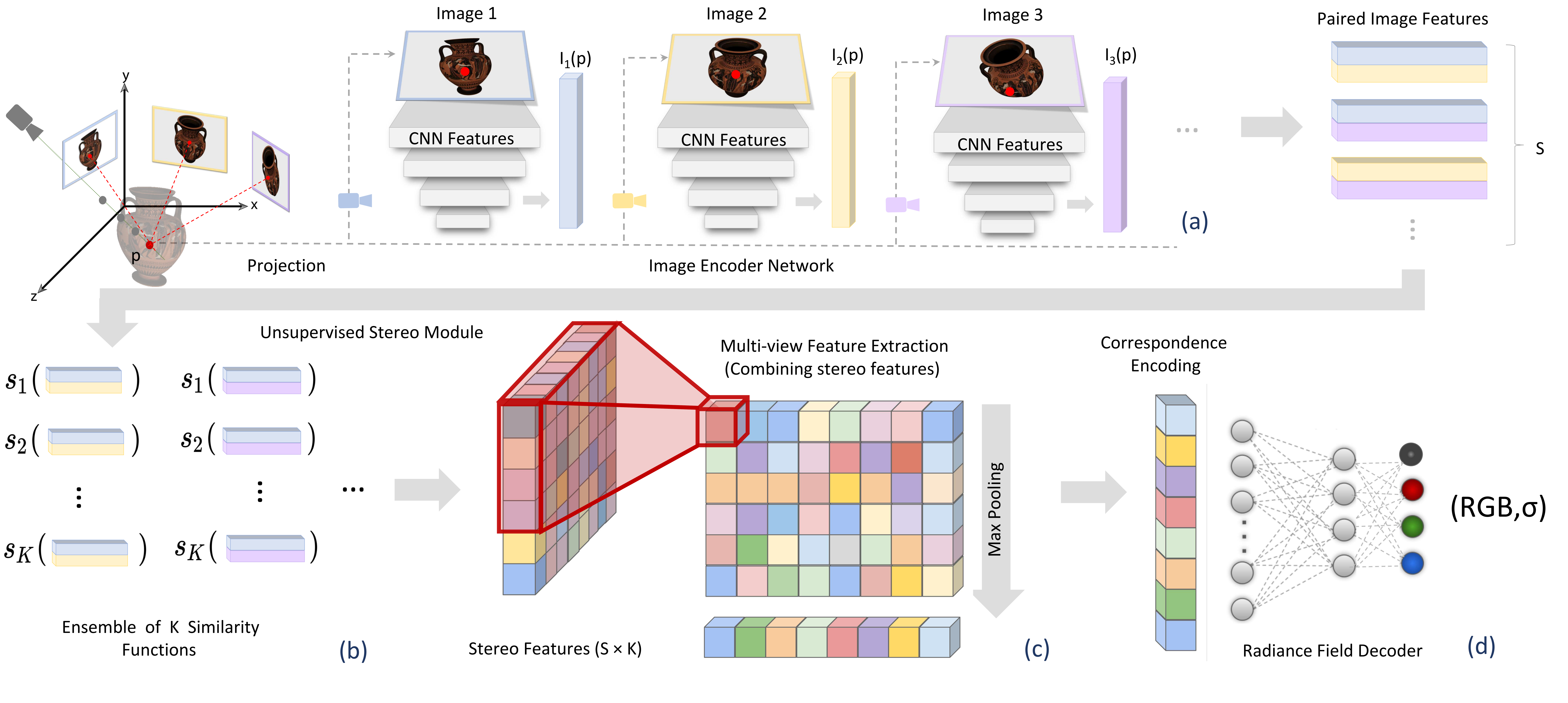}
\caption{\textbf{Our Approach:} For a target view (left camera) we predict RGB color for each pixel. For a pixel, we project a ray into the scene and sample points along it. For a point, $\point \in \mathbb{R}^3$, our goal is to estimate its color, $\pcolor$, and density, $\pdensity$, where density encodes surface regions. (a) First, to encode the location of point $\point$, we project it into each reference view, $\refimg{i}$ and extract features, $\feature{i}$, generated by a 2D CNN at the location of projection. (b) If $\point$ is on a surface and photo-consistent, $\feature{1},\dots,\feature{N}$ will match (see Fig. \ref{fig:method_intuition}).
We emulate the process of finding photo-consistency by applying a learned similarity function $s_k(\cdot,\cdot)$ on all possible combinations. We learn an ensemble of similarity $K$ functions, and obtain a Stereo Features matrix. (c) To aggregate multi-view information beyond pairs, we apply a 2D convolutional CNN to obtain a Multi-view Feature matrix. The matrix is Max Pooled to obtain a compact encoding of correspondence and color, which is decoded by an MLP into color and density (d). Weighted by the density, the color values along the target camera ray are fused into the final pixel color by volume rendering. We train the model end-to-end with image supervision alone.} 
\label{fig:method}
\end{figure*}

\section{Method}
\label{sec:method}

In this section, we present our method \textit{Stereo Radiance Fields} (SRF) for novel view synthesis given sparse and spread-out input views of objects unseen during training. We first give a background in Section~\ref{subsec:methid_generalizing_nerf} and then build SRF on these insights in Section~\ref{subsec:method_srf}.

\subsection{Background}
\label{subsec:methid_generalizing_nerf}

\subsubsection{Generalizing Neural Radiance Fields (NeRF)} To produce color at a pixel of the target view, we shoot a ray from the camera position through the pixel into the scene. We binarize the ray into equal length bins and randomly sample one 3D point within each bin. At each point $\point \in \mathbb{R}^3$ we predict color $\pcolor \in \{0,\dots,255\}^3$ and density $\pdensity \in \mathbb{R}$. Density encodes regions of surface (high where there is surface, low elsewhere). Weighted by the density, the color values along the ray are fused into the final pixel color by volume rendering, following NeRF ~\cite{mildenhall2020nerf}.

NeRF memorizes a scene with a neural function $f$, by learning to output $(\pcolor,\pdensity)$ given spatial location $\point$ and viewing direction $\mathbf{d}$
\begin{equation}
    f_{\mathrm{NeRF}}: (\point, \mathbf{d}) \mapsto (\pcolor,\pdensity).
\end{equation}
This works well for a single scene with dense views. However, it fails to generalize to novel scenes as point coordinates do not carry scene-specific information. The neural network itself becomes the scene representation (Fig. \ref{fig:concerns}-(a)).
Instead, we aim to learn a neural model which emulates multi-view stereo reconstruction and synthesis internally and is conditioned on the scene itself at test time (Fig. \ref{fig:concerns}-(b)). For this, we use a completely different point encoding architecture, which is not scene-specific

\begin{equation} \label{eq:prediction}
    f: (\imageset,\point) \mapsto (\pcolor,\pdensity),
\end{equation}where $\imageset = \{ I_i \}_{i=1}^N$ is the set of $N$ reference images, $\refimg{i}$, with known camera parameters. The design of $f$ is inspired by classical stereo and is explained in Sec.~\ref{subsec:method_srf}.
Note that we do not consider view dependent effects and leave it for future work. This allows us to focus on generalization to novel scenes and sparse inputs.

\subsubsection{Classical Multi-View Stereo}
\label{subsubsec:method_mvs}
Key to classical stereo imaging approaches (Structure-from-Motion, Multi-View Stereo) and our method is the following observation:
In absence of occlusion, surface 3D points of an object project to \emph{corresponding} photo-metrically consistent image regions in the multiple-views, whereas non-surface 3D points land on non-corresponding different regions (Fig.~\ref{fig:method_intuition}). We can invert this observation to find surfaces from images: we can find corresponding regions across views and triangulate them to find a 3D surface point. In classical works, this is done in non-differentiable, multi-step engineered pipelines. First, informative, distinctive regions of \emph{interest} are found. Subsequently, a feature \emph{descriptor} at the interest point is created from local image features, c.f. SIFT~\cite{lowe2004distinctive}. The descriptors from multiple images are matched based on a similarity measure. SRF internally mimics correspondence matching in an end-to-end unsupervised manner (based only on the rendering loss). Our point feature descriptors are learned by a 2D CNN image encoder network. Classical correspondence finding is emulated in SRF by processing point descriptors in pairs.

\subsection{Stereo Radiance Fields (SRF)}
\label{subsec:method_srf}
SRF predicts color and density at a point, $\point$, in 3D space given, 
$\imageset = \{ I_i \}_{i=1}^N$, a set of $N$ reference images, $\refimg{i}$, with known camera parameters.
We structure \textit{SRF}, $f$, in analogy to classical multi-view stereo approaches: (1) To encode the location of point $\point$, we project it into each reference view, $\refimg{i}$, and build a local feature descriptor, $\feature{i}$, (Sec. \ref{subsec:method_encoder}); (2) If $\point$ is on a surface and photo-consistent, $\feature{1},\dots,\feature{N}$ should match (Fig. \ref{fig:method_intuition}); Feature matching is emulated with a learned function, $g_\mathrm{stereo}$, encoding the features from all reference views (Sec. \ref{subsec:method_correspondence_net}); (3) The encoding is decoded by a learned decoder, $\mathrm{dec}$, into the NeRF~\cite{mildenhall2020nerf} representation (Sec. \ref{subsec:method_decoder}). Formally, this decomposes \textit{SRF} to:

\begin{equation}
    f(\imageset,\point) = \mathrm{dec}( g_\mathrm{stereo}(\feature{1},\dots,\feature{N})) \mapsto (\pcolor,\pdensity).
\end{equation}

Figure \ref{fig:method} gives an overview of our method.

\subsubsection{Image Encoder Network}
\label{subsec:method_encoder}
In contrast to NeRF, where the input are point coordinates without scene-specific information, we condition our prediction on the reference images. 
We achieve this by projecting $\point$ into each reference view, $\refimg{i}$, and build a local feature descriptor, $\feature{i}$. For this, we first encode each complete reference image with a shared 2D CNN. We build $\feature{i}$ by extracting the deep features from each CNN layer at the location of the points $\point$ projection. This makes $\feature{i}$ a multi-scale feature descriptor, as 2D CNNs naturally encode local information in their first layers up to global information in later layers with a high receptive field (Fig.~\ref{fig:method}-(a) ``Image Encoder Network'').
Because the point projection is in continuous space, whereas the features are in a discrete grid, we use bilinear interpolation for extraction. When $\point$ projects outside of an image we use zero padding. See appendix for further details.

\subsubsection{Unsupervised Stereo Module}
\label{subsec:method_correspondence_net}
We build on the intuition of Multi-View Stereo: when a 3D point $\point$ is projected to photo-metrically consistent regions, $\point$ is likely to lie on a surface, and hence a high density $\pdensity$ should be predicted. 
In order, to process an arbitrary number of views, the stereo module processes feature descriptors of views in \emph{pairs}.  
Specifically, we aim at learning mappings of feature pairs $\feature{i},\feature{j}$:
\begin{equation}
    s: (\feature{i},\feature{j}) \mapsto x \in \mathbb{R}_0^+,
\end{equation}
that allow the network to learn image scores useful for correspondence finding or propagate image color. 
Note, although our formulation is based on pairwise processing analog to similarity computation, \emph{correspondences are not explicitly computed}.
We represent each mapping $s$ in the network using a single neuron.
In practice, each possible pair $(\feature{i},\feature{j})$ with $i,j \in 1,\dots,N, i \neq j$ is input to a neuron with ReLU non-linearity to ensure non-negative outputs (Fig. \ref{fig:method}-(b)). This yields a vector $\mat{x}$ of size $S=N^2-N$ with one entry per pair. Instead of relying on only a single neuron, we apply a bank of neurons, $s_k(\cdot,\cdot), \ k=1 \hdots K$ in the same fashion. Each neuron might learn different similarities, or specialize in propagating color.
We concatenate the output vector $\mat{x}_k$ of each neuron in the bank into a \emph{Stereo Feature} matrix $\mat{X} = [\mat{x}_1\hdots \mat{x}_K] \in \mathbb{R}^{S\times{K}}$ whose height is the number of pairs $S$ and the width is the number of neurons $K$ used (Fig.~\ref{fig:method}-(b) ``Stereo Features''). The Stereo Feature matrix can be efficiently computed by arranging feature pairs in a matrix and convolving it with the neuron bank.

Pairwise photo-consistency is, however, not a sufficient condition to identify surface points. 3D points might project to photo-consistent image regions in a stereo pair when reference views are captured nearby but not on a third view. We aggregate information from multi views by convolving the Stereo Feature matrix along the direction of pairs of views. Specifically, we aggregate $4$ pairs in the height direction and all similarity measures along the width direction (Fig.~\ref{fig:method}-(c) ``Multi-view Feature Extraction'').

To merge view-pair information into a single vector $\mathbf{y} \in \mathbb{R}^K$, we run Max Pooling in the direction of views. Note that the complete stereo module by design, is flexible for varying number of input views during training and testing: the Max Pooling step is computing a vector $\mathbf{y}$ of fixed dimension $K$ given varying number of input views. This constitutes the unsupervised Stereo Module, denoted by 
\begin{equation}
    g_\mathrm{stereo}: (\feature{1} ,\dots,\feature{N}) \mapsto \mathbf{y} \in \mathbb{R}^K.
\end{equation}

\subsubsection{Radiance Field Decoder}
\label{subsec:method_decoder}

The last stage of our network is to decode the stereo encoding $\mathbf{y} = g_\mathrm{stereo}(\feature{1} ,\dots,\feature{N})$ of point $\point$ into the final color $\pcolor$ and density $\pdensity$. For this, we rely on a simple MLP network denoted by 
    \begin{equation}
    \mathrm{dec}: \mathbf{y} \mapsto (\pcolor, \pdensity).
\end{equation}

Sampled colors along a ray are fused based on their density following volume rendering~\cite{Max1995volumerendering,mildenhall2020nerf}.
The training of the network is done fully end-to-end using only multi-view images without 3D data or supervision on the stereo module (Fig.~\ref{fig:method}-(d)). We use the $L2$ loss for comparing the rendered prediction with the target image. Please consider the appendix for further architectural details.

%% file: sections/Experiments.tex
\begin{figure*}[t]
\centering
\includegraphics[width=1.0\linewidth]{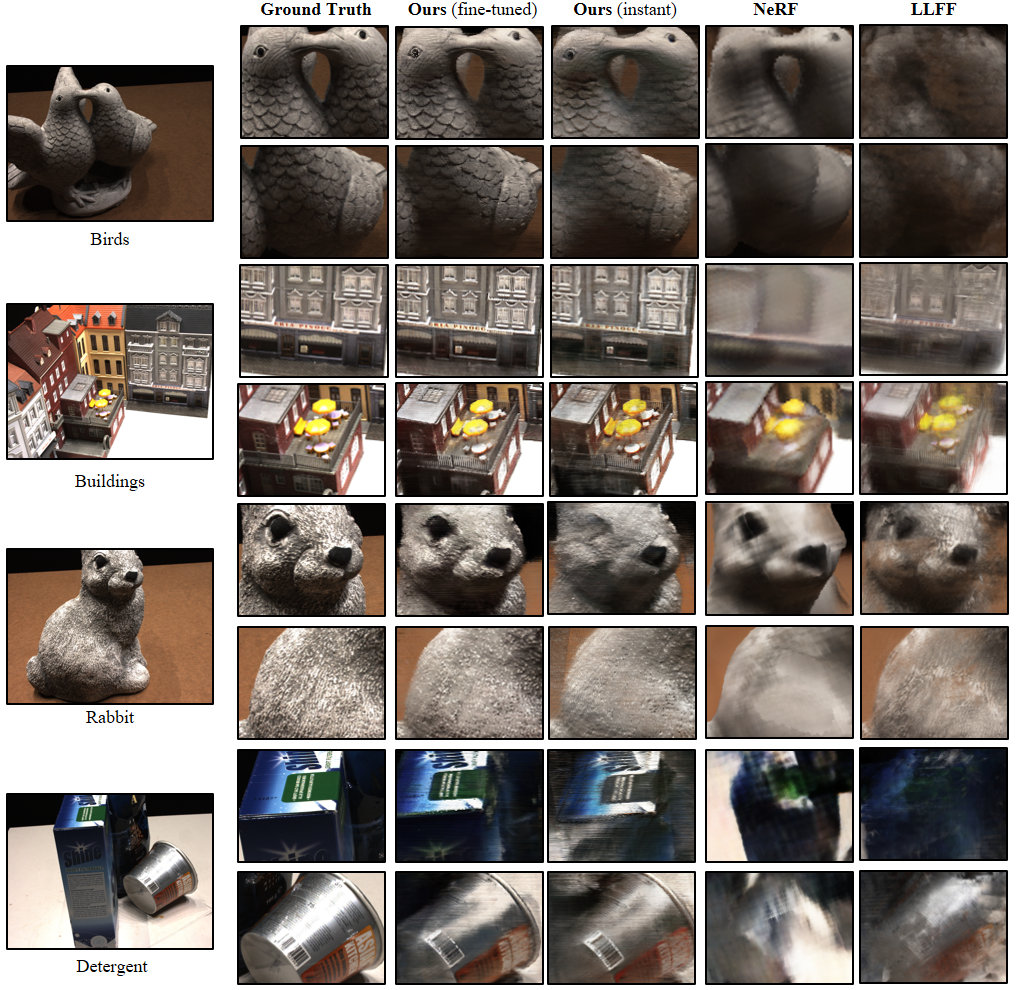}
\caption{\textbf{Comparisons:} We compare our method on test views of scenes of DTU. Given $10$ reference view images of a novel scene at test time. Our method infers sharp and detailed objects in both appearance and geometry, such as the feathers and eyes of \textbf{birds}, the letters and small benches in \textbf{buildings}, the texture of the \textbf{rabbit}, and the logo of a \textbf{detergent}. \textit{Rabbit} and \textit{detergent} scenes benefit most from fine-tuning. NeRF finds approximate, smooth geometry and yields blurry textures for \textit{birds}, \textit{buildings} and \textit{rabbit}. For the detergent scene, it struggles to generate consistent geometry or appearance. LLFF creates some sharp image regions at the letters of \textit{buildings} and the texture of the \textit{rabbit} but results are usually overlaid with strong blending and ghosting effects.}

\label{fig:exps_comparisons}
\end{figure*}

\section{Experiments}
\label{sec:experiments}

\begin{table*}
	\footnotesize
\begin{center}
\begin{tabular}{l|c|c|c|c|c|c|c|c|c|c|c|c}

 & \multicolumn{3}{c|}{Birds }   & \multicolumn{3}{c|}{Buildings }  & \multicolumn{3}{c|}{Rabbit } & \multicolumn{3}{c}{Detergent } \\
  & \multicolumn{1}{c}{PSNR $\uparrow$}  &\multicolumn{1}{c}{SSIM  $\uparrow$}&\multicolumn{1}{c|}{LPIPS  $\downarrow$} & \multicolumn{1}{c}{PSNR $\uparrow$}  &\multicolumn{1}{c}{SSIM  $\uparrow$}&\multicolumn{1}{c|}{LPIPS  $\downarrow$} & \multicolumn{1}{c}{PSNR $\uparrow$}  &\multicolumn{1}{c}{SSIM  $\uparrow$}&\multicolumn{1}{c|}{LPIPS  $\downarrow$}& \multicolumn{1}{c}{PSNR $\uparrow$}  &\multicolumn{1}{c}{SSIM  $\uparrow$}&\multicolumn{1}{c}{LPIPS  $\downarrow$} \\\hline \hline


LLFF   
& 18.65 & 0.51 & 0.44
& 15.13 &  0.39 &  0.40
&  17.59 & 0.41 &  0.49
& 14.73 &  0.49 &  0.48    \\

NeRF   
&  15.09 & 0.29   &  0.71   
&   17.68  & 0.51  &       0.33  
&   18.24 &     0.40     & 0.59 
&     9.73 &     0.32     & 0.64    \\

Ours
& 23.36 & 0.65 & 0.35
& 17.22 &  0.57 &  0.29
& \textbf{18.79} &  0.48 &  0.49 
& 16.75 & 0.48 &  0.48      \\

Ours(ft)
&  \textbf{24.97} & \textbf{0.72}   &  \textbf{0.27}   
&  \textbf{19.71} & \textbf{0.70}  &  \textbf{0.18}  
&  18.06 & \textbf{0.55}   &  \textbf{0.40}   
&  \textbf{16.97} & \textbf{0.60}   &   \textbf{0.37}      \\
\end{tabular}

\end{center}
\vspace{-1mm}
\caption{\textbf{Quantitative Results:} Quantitative results on DTU dataset, reported in PSNR, SSIM (higher is better) and LPIPS~\cite{zhang2018perceptual} (lower is better). Ours(ft) indicates fine-tuning. We outperform all baselines consistently. SRF without fine-tuning already outperforms baselines, fine-tuned SRF produces even sharper geometry, appearance, and far fewer artifacts than all baselines.}
\label{tab:exps_quantitative}
\vspace{-3mm}
\end{table*}

We, first, study the generalization ability of SRF when trained on a variety of generic objects and scenes. In Sec. \ref{subsec:exps_generalization}, we observe that our model indeed learns generalizing structure applicable on novel scenes, given only a sparse number of views that are arbitrarily spread-out. Furthermore, we find that our model can produce 3D colored meshes from only 10 views, despite being trained for a view synthesis task as shown in Sec. \ref{subsec:exps_meshing}. These observations suggest that incorporating \textit{geometry and data} helps generalization. Finally, in Sec. \ref{subseq:exps_extreme_generalzation}, we show that the multi-view structure of SRF naturally generalizes, even when learned on a \textit{single} object for a \textit{limited time}. 

\paragraph{Data.} We conduct our experiments on the publicly available DTU Multi-View Stereopsis Dataset~\cite{aanaes2016dtumvs}. It consists of 124 different scenes, including very diverse objects (E.g., buildings, statues, groceries, fruits, bricks, etc.). We split the scenes into test, validation, and training splits (see appendix for more details). We randomly sample 10 images of a scene as input to SRF. For evaluation and training purposes, we sample a different view as the target view.


\begin{figure}
\centering
\footnotesize
   \includegraphics[width=1\linewidth]{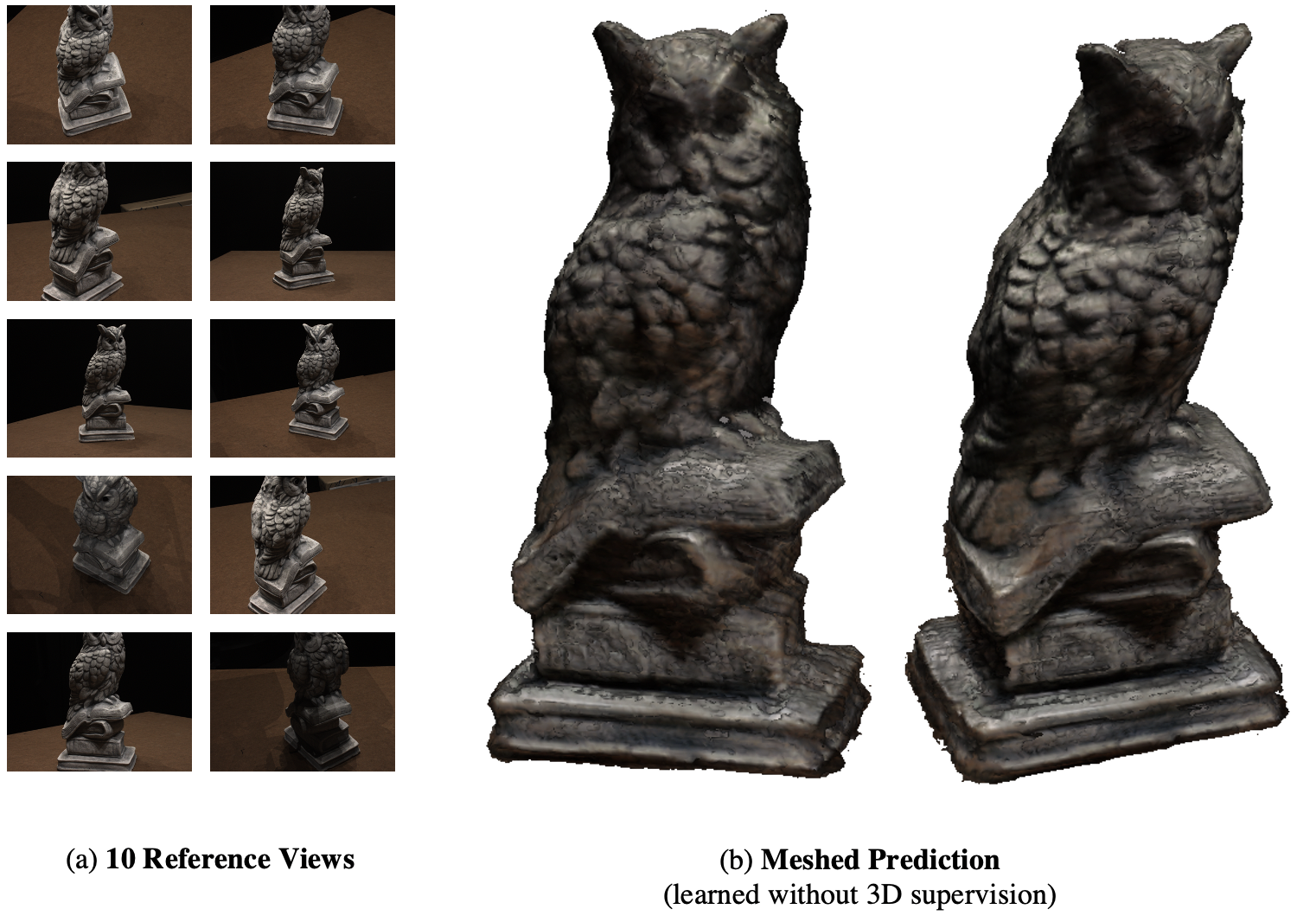}
\caption{\textbf{Meshing Predictions.} Given only 10 images of a scene, SRF can produce colored meshes from the resulting density. We posit that SRF implicitly learns 3D reconstruction and view synthesis jointly from only 10 views even when no 3D supervision was provided during training.}
\label{fig:exps_3D_reconstruction}
\vspace{-3mm}
\end{figure}

\paragraph{Baselines} 
We contrast our approach with  \textbf{NeRF}~\cite{mildenhall2020nerf}. NeRF requires scene-specific optimization. We use publicly available code to train NeRF models for each scene using $10$ input images. Training a scene-specific model took 2 days. Once trained, novel views can be synthesized. 
We also compare to an off-the-shelf publicly available \textbf{LLFF}~\cite{mildenhall2019local} model. Like ours, LLFF allows for generalization to test scenes\footnote{We did not have access to the training code of LLFF. Therefore, we use an off-the-shelf model provided by the authors. It is possible that results may improve by fine-tuning the LLFF model on DTU dataset.}.  Instead of a continuous 3D representation, reference images are sliced into multiple depth layers. For synthesis of a target view, neighboring reference images are warped into the target view and blended together.

\begin{figure*}[t]
\centering
\includegraphics[width=1.0\linewidth]{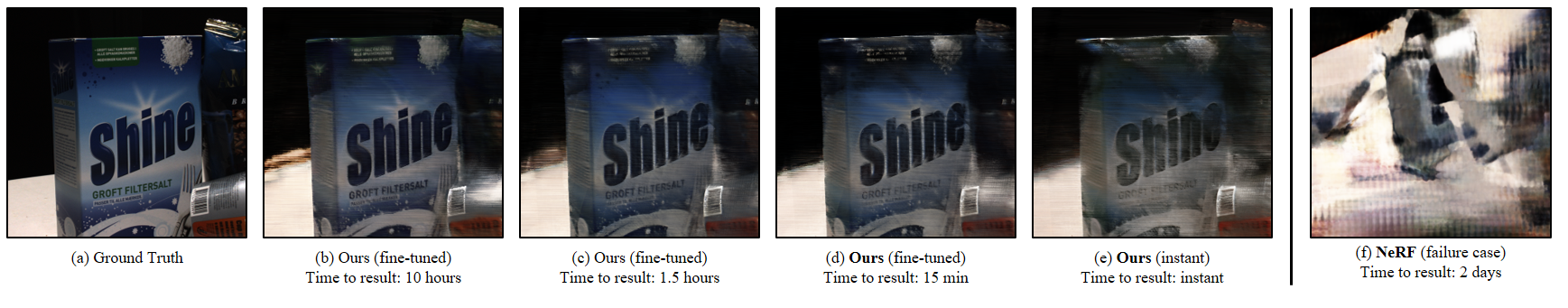}
\caption{\textbf{Effect of Fine-Tuning our Method.} 
While pure NeRF struggles here \textbf{(f)}, SRF generates a reasonable result \textbf{(e)}. We further improve the result by fine-tuning with the test images. We observe around 15 minutes to be a good trade-off between quality and speed, and that around 90 minutes results converge. Not only this results in sharper results as compared to the baselines, but it also reduces optimization time from 2 days to minutes. 
}
\label{fig:exps_finetuning_failiure}
\end{figure*}

\subsection{Unconstrained Generalization}
\label{subsec:exps_generalization}

In this experiment, we target to learn a model that is able to perform novel view synthesis on any unseen test scene. For this, we sample a random train (109 scenes), test (10 scenes), and validation (5 scenes) split of the full DTU dataset (see appendix). We train our method until validation minimum is reached for around 3 days on a single NVIDIA Quadro RTX 8000.

Given only 10 views of a novel scene at test time, our method is able to create sharp objects in the rendered novel views and outperforms baselines. We show qualitative analysis in Figure~\ref{fig:exps_comparisons} and quantitative analysis in Table \ref{tab:exps_quantitative}. Our approach generalizes to new scenes instantly and can operate on sparse and arbitrarily spread-out multi-views. Each NeRF model takes 2 days for the scene-specific optimization. 
Instead, our SRF can be learned from many scenes, thanks to the architecture which emulates geometric stereo matching.
We find this to be key for novel view synthesis from sparse data. Moreover, we can enrich the geometric and learning concept by the idea of optimized scene representations. For this, we fine-tune our model for a short period of time. We show the effect of fine-tuning our method in Fig. \ref{fig:exps_finetuning_failiure} (b)-(d) and also in Fig. \ref{fig:exps_comparisons}. Around 15 minutes yields a good trade-off between quality and speed and at around 90 minutes fine-tuning converges. Not only does this result in sharper results as compared to the baselines, but the optimization time is also reduced from multiple days to minutes. 
We observe that a NeRF model trained on the sparse and spread-out views may also lead to degenerate results as shown in Fig. \ref{fig:exps_finetuning_failiure}-(f). We refer the reader to the appendix for more details.

\begin{figure}
\centering
\footnotesize
   \includegraphics[width=1\linewidth]{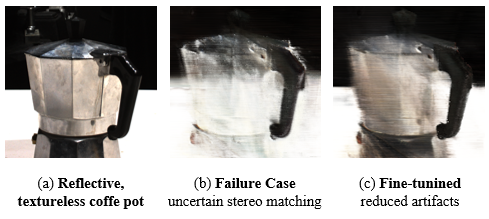}
\caption{\textbf{Limitation} Our neural architecture of SRF is strongly inspired by classical stereo matching. Modeling reflections and texture-less regions is challenging. Fine-tuning SRF ameliorate this issue, though does not totally overcome it. }
\label{fig:limitation}
\vspace{-3mm}
\end{figure}

Finally, we find that challenging BRDFs and reflective regions can pose problems for our method based on stereo matching. We observe that fine-tuning helps mitigate some issues (Fig. \ref{fig:limitation}). Introducing view-dependent modeling into SRF, which we dropped in Eq. \ref{eq:prediction}, may likely solve this issue and is an interesting future work direction.

\begin{figure}
\centering
\footnotesize
   \includegraphics[width=1\linewidth]{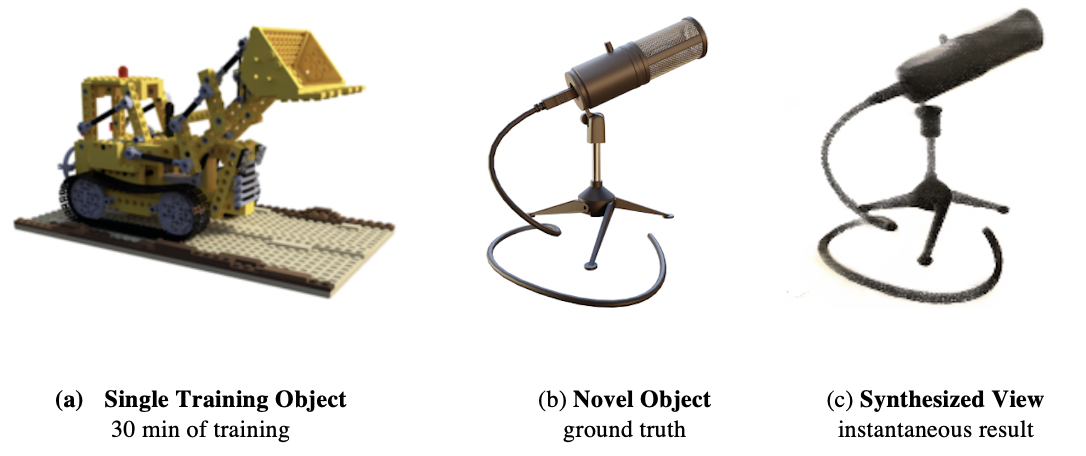}
\caption{\textbf{Natural Generalization Capability.} We train SRF only on a single object, a tractor, for as little as \textit{30 min}, and apply it without fine-tuning to a microphone. It is apparent that geometry and some color generalize even in this extreme setting, despite the large differences in geometry and appearance between the tractor and the microphone.
We attribute this to the classical stereo geometry build into our network by design.}
\label{fig:exps_natural_generalization}
\vspace{-3mm}
\end{figure}

\subsection{Meshing Predictions}
\label{subsec:exps_meshing}
In order to mesh the prediction, we evaluate SRF conditioned on 10 images in a dense grid of points enclosing the objects. SRF predicts color and density for each point. We then threshold the density at the grid and run Marching Cubes~\cite{Lorensen1987marching_cubes} to obtain a mesh. For each vertex we find on the mesh, we take its coordinates and input them to the SRF to predict color and add it to the mesh. See Fig. \ref{fig:exps_3D_reconstruction} for a result.

\subsection{Natural Generalization Capability}
\label{subseq:exps_extreme_generalzation}

Previously, we found that incorporating \textit{geometry and data} helps generalization. Next, we validate that our architecture naturally generalizes by design. We take a radical setup for this: we train on a single object (a synthetic tractor~\cite{mildenhall2020nerf}) for as little as 30 minutes and inspect novel view synthesis of a very different object (a microphone from NeRF data). We observe generalization despite large differences in appearance and geometry as shown in Fig. \ref{fig:exps_natural_generalization}.

%% file: sections/Conclusions.tex
\section{Discussion and Conclusion}
\label{sec:conclusion}
We introduced Stereo Radiance Fields, a neural view synthesis model designed to emulate components of classical multi-view stereo.
Instead of predicting radiance and color based on point-direction coordinates, we project each 3D point to multiple views, extract features, and process them in \emph{pairs}.
This learns an ensemble of scores driven only by a self-supervised rendering loss, that allow for computation of implicit correspondences. The process emulates feature matching in classical stereo within an end-to-end learnable network for view synthesis.

Experiments demonstrate that SRF learns common structure across multiple scenes. 
We train a SRF model on multiple scenes from the DTU dataset, and show that SRF \emph{generalizes}, producing realistic images. 
Furthermore, in contrast to prior work which requires dense views, we use arbitrarily \emph{sparse} spread-out $10$ views as input. 
We show that results further improve after $10$-$15$ minutes of fine-tuning on these target $10$ views. Remarkably, in the sparse view setting ($10$ views), our approach significantly outperforms the SOTA methods, even when we train them on the new scene for $2$ days. Finally, we show that SRF implicitly compute an interpretable 3D representation allowing for colored meshing -- without using 3D supervision.

In summary, SRF builds on classical multi-view stereo and recent neural rendering ideas but combines them in a unified end-to-end learnable architecture. 
We think the interplay of classical geometric computer vision with neural rendering is an exciting avenue, which deserves further exploration. Future work may extend them to model challenging BRDFs, and 4D space-time view synthesis of dynamic scenes from in-the-wild samples that are inherently sparse.

%% file: sections/Acknowledgements.tex
\hyphenation{For-schungs-ge-mein-schaft}
We thank the RVH group for their feedback. This work is funded by the Deutsche Forschungsgemeinschaft (DFG, German Research Foundation) - 409792180 (Emmy Noether Programme, project: Real Virtual Humans).